\documentclass[sn-mathphys,Numbered]{sn-jnl}

\usepackage{graphicx}%
\usepackage{multirow}%
\usepackage{amsmath,amssymb,amsfonts}%
\usepackage{amsthm}%
\usepackage{mathrsfs}%
\usepackage[title]{appendix}%
\usepackage[dvipsnames]{xcolor}%
\usepackage{textcomp}%
\usepackage{manyfoot}%
\usepackage{booktabs}%
\usepackage{algorithm}%
\usepackage{algorithmicx}%
\usepackage{algpseudocode}%
\usepackage{listings}%
\usepackage{subcaption}%
\usepackage{tikz}%
\usepackage{pgfplots}%
\usepackage{adjustbox}

\pgfplotsset{every axis x label/.style={
  at={(0.5,0)},
  below,
  yshift=-5pt}}
\pgfplotsset{every axis y label/.style={
  at={(0,0.5)},
  xshift=-15pt,
  rotate=90}}
\pgfplotsset{compat=1.18}

\theoremstyle{thmstyleone}%
\theoremstyle{thmstyletwo}%

\theoremstyle{thmstylethree}%

\newtheorem{theorem*}{Theorem}[section]
\newtheorem*{proposition*}{Proposition}

\begin{document}

\abstract{Scientific discovery and engineering design are currently limited by the time and cost of physical experiments, selected mostly through trial-and-error and intuition that require deep domain expertise. Numerical simulations present an alternative to physical experiments but are usually infeasible for complex real-world domains due to the computational requirements of existing numerical methods.
Artificial intelligence (AI) presents a potential paradigm shift by developing fast data-driven surrogate models. 
In particular, an AI framework, known as Neural Operators, presents a principled framework for learning mappings between functions defined on continuous domains, e.g., spatiotemporal processes and partial differential equations (PDE). 
They can extrapolate and predict solutions at new locations unseen during training, i.e., perform zero-shot super-resolution. 
Neural Operators can augment or even replace existing simulators in many applications, such as computational fluid dynamics, weather forecasting, and material modeling, while being 4-5 orders of magnitude faster. 
Further, Neural Operators can be integrated with physics and other domain constraints enforced at finer resolutions to obtain high-fidelity solutions and good generalization. Since Neural Operators are differentiable, they can directly optimize parameters for inverse design and other inverse problems.
We believe that Neural Operators present a transformative approach to simulation and design, enabling rapid research and development.}


\title[Neural Operators]{Neural Operators for Accelerating Scientific Simulations and Design}


\author[1]{ \sur{Kamyar Azizzadenesheli}}\email{kamyara@nvidia.com}

\author[1]{\fnm{Nikola Kovachki} }\email{nkovachki@nvidia.com}

\author[2]{\fnm{Zongyi Li} }\email{zongyili@caltech.edu}
 
\author[2]{\fnm{Miguel Liu-Schiaffini} }\email{mliuschi@caltech.edu}

\author[1]{\fnm{Jean Kossaifi} }\email{jkossaifi@nvidia.com}

\author[1,2]{\fnm{Anima Anandkumar} }\email{anima@caltech.edu}

\affil[1]{ \orgname{NVIDIA}, \orgaddress{\street{2701 San Tomas Expressway}, \city{ Santa Clara}, \postcode{95050}, \state{CA}, \country{USA}}}

\affil[2]{ \orgname{Caltech}, \orgaddress{\street{1200 E. California St}, \city{Pasadena}, \postcode{91125}, \state{CA}, \country{USA}}}

\maketitle

\section{Introduction}\label{sec:intro}

The scientific method has been the cornerstone of research and development. It consists of a set of hypotheses formed through inductive reasoning based on the observations collected, which are first systematically tested, before being iteratively updated. However, testing the hypotheses in many scientific domains involves slow and expensive real-world experiments, which limits scientists to a small set of hypotheses that are driven by prior knowledge and intuition.  If expensive experiments can be replaced with faithful computational simulations, we can vastly expand our hypothesis space for exploration. 

In principle, it is possible to create such desired simulations since we know the (approximate) governing laws of many physical, chemical, and engineering processes. The physical world around us is often modeled by governing partial differential equations (PDEs) derived from first principles, e.g., laws of thermodynamics, chemical interactions, and mechanics~\cite{evans2022partial}. For example, the aerodynamics of cars and planes can be modeled using the Navier-Stokes PDE defined on the geometry of the surface~\cite{batchelor1967introduction}. However, it is currently infeasible to run numerical simulations at the scale and fidelity needed to replace physical experimentation entirely. This is because existing numerical methods used in simulations require a fine computational grid to guarantee convergence of the iterations.

Climate modeling, for example, uses PDEs to simulate the future of Earth's climate and is critical for policy-making and developing strategies for climate change mitigation. In this case, physical experimentation is impossible, and simulations are the only option. However, current numerical methods require massive computing power
to resolve the physics at finer scales where the effects of storms and cloud turbulence can be faithfully reproduced, and even the world's most powerful supercomputers do not suffice~\cite{schneider2017climate}. Moreover, numerical methods have other shortcomings pertinent to applications involving observational data where the governing laws are only partially specified or inaccurate. Current numerical methods cannot directly exploit observational data for such cases since they are not explicitly data-driven. Additionally, numerical methods are not generally differentiable, thus making inverse problems and design optimization expensive and slow \cite{tarantola2004inverse}.

Artificial intelligence (AI) is now transforming our ability to create fast, high-fidelity simulations of many complex multi-scale processes~\citep{brunton2020machine, kochkov2021machine}. Previous developments such as sparse representation \citep{brunton2016discovering}, recurrent neural networks \citep{vlachas2018data}, reservoir computing \citep{pathak2018hybrid} and others have shown promise in modeling dynamical systems.
The latest advances in AI, termed Neural Operators~\citep{li2020fourier,kovachki2023neural}, can model such phenomena and capture finer scales, leading to high-fidelity solutions. We have shown that Neural Operators enjoy several advantages: 
(1) Neural Operators can serve as fast and accurate surrogates for numerical methods and overcome the high computational requirements of numerical methods. Unlike conventional numerical methods, Neural Operators can learn an efficient feature representation or basis that does not require a fine grid. Further, once the Neural Operator model is trained, the inference is typically a single forward pass, as opposed to a large number of iterations required in numerical methods. In addition,  since Neural Operators are AI models, they can build on the latest hardware and software advances of accelerated computing~\cite{dally2021evolution} to obtain the best possible speedups with the learned representations. This has resulted in  $45000\times$ speedup in the weather forecast~\citep{pathak2022fourcastnet}, $26000\times$ in the automotive industry~\citep{ginopaper}, and $700,000\times$ carbon dioxide storage~\citep{wen2023real}, along with many other scientific computing domains.
(2) Neural Operators can learn directly from observational data and do not suffer from modeling errors present in numerical methods that rely solely on simplified equations for many applications. 
(3) Neural Operators are differentiable and can be used directly for inverse problems such as optimized design generation. In contrast, traditional solvers are usually not differentiable and require a large number of iterations using methods such as Markov Chain Monte-Carlo (MCMC). 
(4) Neural Operators are also democratizing science since running the trained Neural Operators does not require the deep domain expertise, needed to set up and run many traditional solvers. Moreover, many applications, such as weather forecasting, that previously required large CPU clusters can now be carried out using consumer desktop GPUs and open-source models, enabling widespread accessibility~\cite{pathak2022fourcastnet,bonev2023spherical,url:fcn-ecmwf}.

\section{Neural Operators}
\begin{figure}
    \centering
    \begin{subfigure}{0.45\textwidth}
    \includegraphics[width=0.84\textwidth]{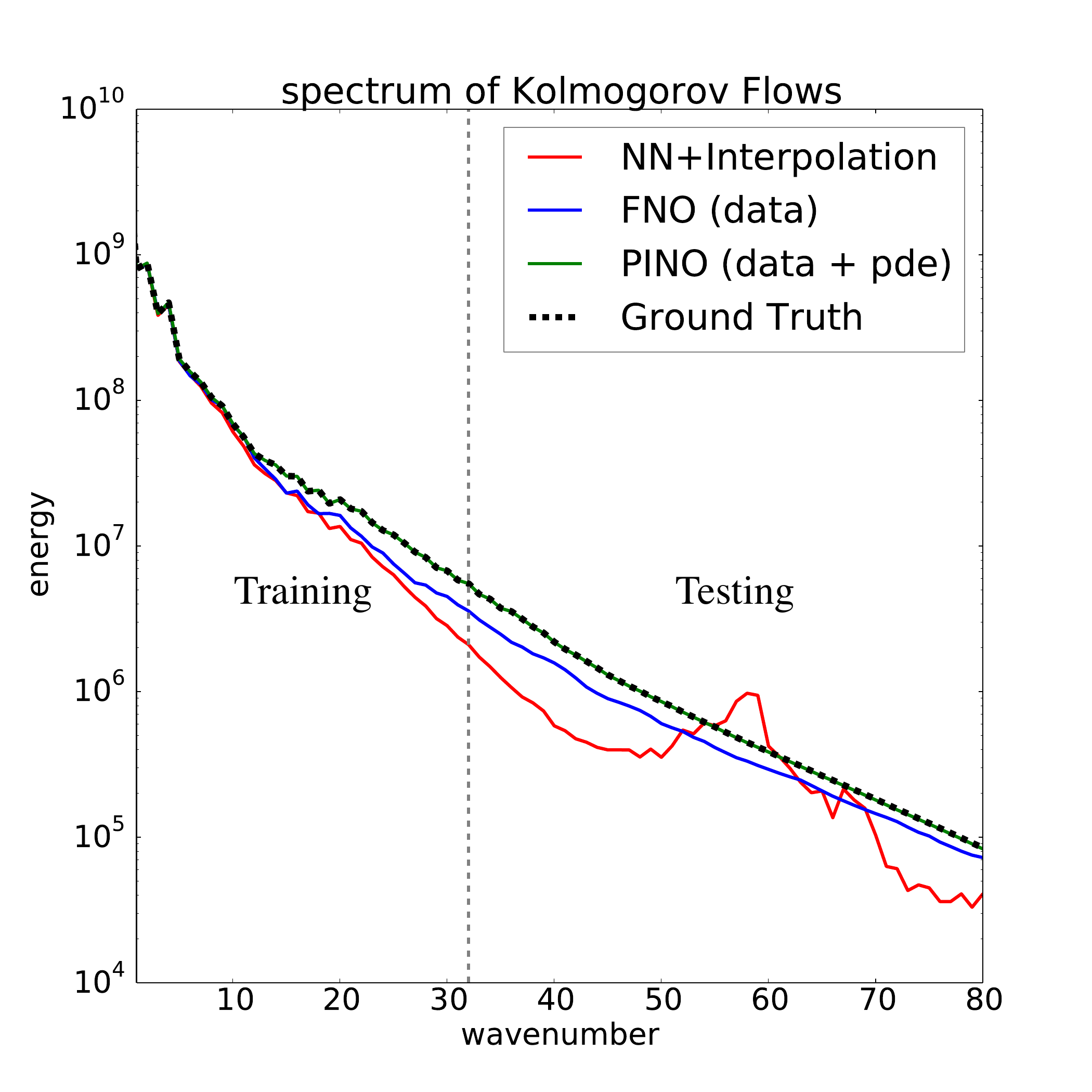}
    \end{subfigure}
    \begin{subfigure}{0.45\textwidth}
    \includegraphics[width=0.9\textwidth]{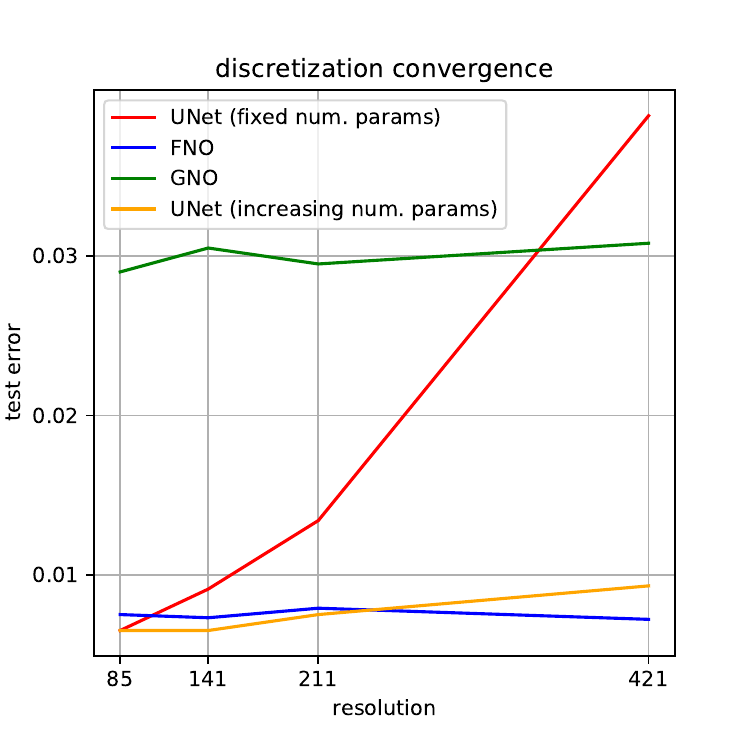}
    \end{subfigure}
    \caption{{\bf Left: }The x-axis is the Fourier wavenumber and y-axis is the energy per spectrum. Fourier Neural Operators (FNO) can extrapolate to unseen frequencies in Kolmogorov Flows \cite{pino} using only limited resolution training data. Physics-informed Neural Operator (PINO) uses both training data and the PDE equation for the loss function, and can perfectly recover the ground-truth spectrum.  A trained UNet with trilinear interpolation (NN+Interpolation) has severe distortions at higher frequencies, beyond the resolution of training data. 
    {\bf Right:} The x-axis is the resolution of the test data, and y-axis is the test error at that given resolution.  Neural Operators are discretization convergent, meaning the model converges to the target continuum operator as the discretization is refined. On the Darcy equation, we train each fixed architecture UNet, FNO, and Graph Neural Operator (GNO) at a given resolution and test at that same resolution \cite{kovachki2023neural} (no super-resolution).
    As shown in the figure, FNO and GNO have consistent errors as resolution increases, but UNet has increasing errors since the size of its receptive field changes with resolution, and it does not enjoy the guarantees of discretization convergence. UNet is only able to maintain the same test error as resolution increases if the number of parameters increases. In this case, we increase the convolutional filter size, corresponding to about 2.2M, 6.0M, 11.8M, and 19.3M parameters, respectively.
    }
    \label{fig:spectrum}
\end{figure}

\subsection{Motivation}

AI models such as transformers and convolutional neural networks (CNN) have shown impressive performance in domains such as text and images~\cite{NIPS2017_3f5ee243}. However, in our opinion, standard neural networks have fundamental limitations that make them unsuitable for scientific modeling. They have limited generalization beyond conditions seen during training. In particular, they are limited to learning and making predictions at the resolution of the training data. This is a significant limitation since many scientific phenomena occur on continuous domains, e.g., PDEs such as fluid flows, wave propagation, and material deformation, even though only discretized observations are available for training. In contrast, traditional numerical methods to solve PDEs can predict solutions at any location on the given continuous domain. Furthermore, in many applications, the availability of high-resolution data may be limited, and combining it with larger amounts of low-resolution data during training is not feasible in standard neural networks, which only support fixed-resolution inputs and outputs.

A simple solution to the above issue that enables pre-trained neural networks to predict at any resolution is to use standard interpolation methods, e.g., bilinear interpolation on the predictions made at locations on the training grid. However, such approaches perform poorly on many processes, e.g., fluid flows, since they fail to capture finer scales accurately, as seen in Fig.~\ref{fig:spectrum}. This is because simple interpolation schemes cannot capture finer scales that are not present in limited-resolution training data.  Moreover, standard neural networks are unsuitable for incorporating training data at multiple resolutions, e.g., climate modeling datasets~\cite{haarsma2016high}. Even if modifications are made to incorporate multiple resolutions at inputs or outputs~\cite{yuan2021hrformer}, they are not principled and have no guarantees for capturing finer scales. Neural networks suffer from a fundamental limitation, viz.,  they only learn mappings between inputs and outputs of fixed dimensions. As such, they do not have the expressive power to capture mappings between functions on continuous domains, known as operators. Due to this reason, we believe that standard neural networks are unsuitable for replacing numerical solvers in scientific applications. 

To overcome the above limitation of neural networks, we propose Neural Operators that can predict the solution at any location in the output domain and are not limited to the grid of training data~\citep{li2020neural,kovachki2023neural}. Neural Operators accomplish this by approximating the underlying operator, which is the mapping between the input and output function spaces, each of which can be infinite-dimensional. See Fig.\ref{fig:nnvsno}. In other words, Neural Operators approximate the continuum mapping even when trained only on discrete data, and thus, they have the ability to capture the finer scales more faithfully. 

In Fig.\ref{fig:spectrum}, we show the prediction of a Neural Operator, Fourier Neural Operator (FNO), for fluid dynamics~\citep{li2020fourier}, and contrast it with UNet, a popular neural network trained at a fixed resolution and then augmented with trilinear interpolation. We find that FNO follows the trend of ground-truth energy decay in the frequency domain, even beyond the training frequencies, while UNet with interpolation fails to predict the finer scales. Fig.~\ref{fig:spectrum} demonstrates that empirically, FNOs and other Neural Operators can do zero-shot super-resolution and extrapolate to unseen higher frequencies. In contrast, as discussed in~\citep{pino}, band-limited operators and representation-equivalent operators~\citep{bartolucci2023neural}, such as Spectral Neural Operators~\citep{fanaskov2022spectral}, cannot generate new (higher) frequencies since their representation space is fixed. Therefore, they introduce an irreducible approximation error based on the size of the predefined representation space. Since the goal of operator learning is to find the underlying solution operator in the continuum, we believe discretization convergence is a more useful strategy.
However, the error cannot be expected to be (nearly) zero since the model, during training, cannot access information at the finer scales due to the limited resolution of the samples. In scientific modeling, we usually have access to physics constraints such as PDEs that fully constrain the system or conservation laws and symmetries that partially specify the system. The Physics-Informed Neural Operator (PINO)~\citep{pino} incorporates both data and physics losses, leads to even better extrapolation to higher frequencies, and almost perfectly matches the ground-truth spectrum in Fig.~\ref{fig:spectrum}.

\subsection{Formulation}

\begin{figure}[t]
    \centering
    \begin{subfigure}{0.32\textwidth}
        \centering
        \includegraphics[width=0.9\textwidth]{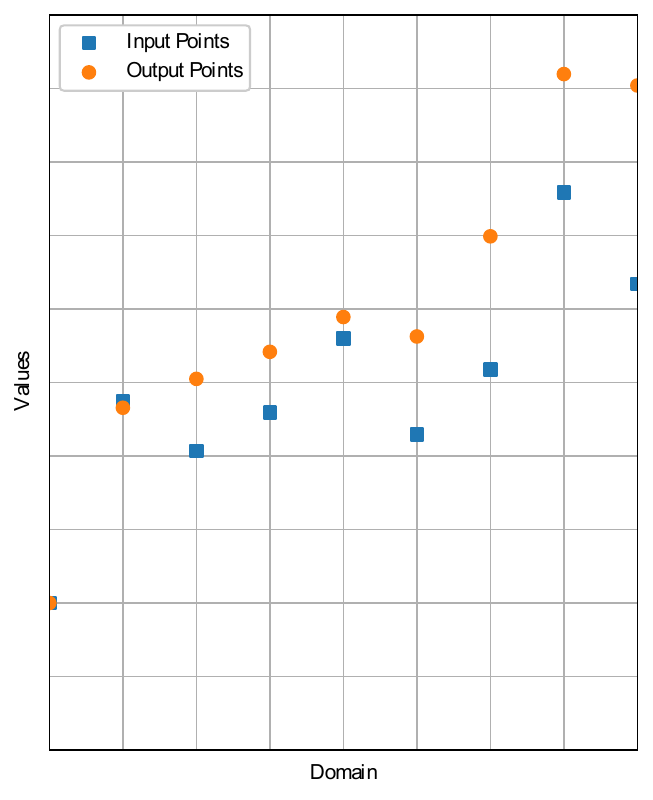}
        \caption{NN learns a mapping between input and output points on a fixed, discrete grid.}
    \end{subfigure}%
    \hfill
    \begin{subfigure}{0.32\textwidth}
        \centering
        \includegraphics[width=0.9\textwidth]{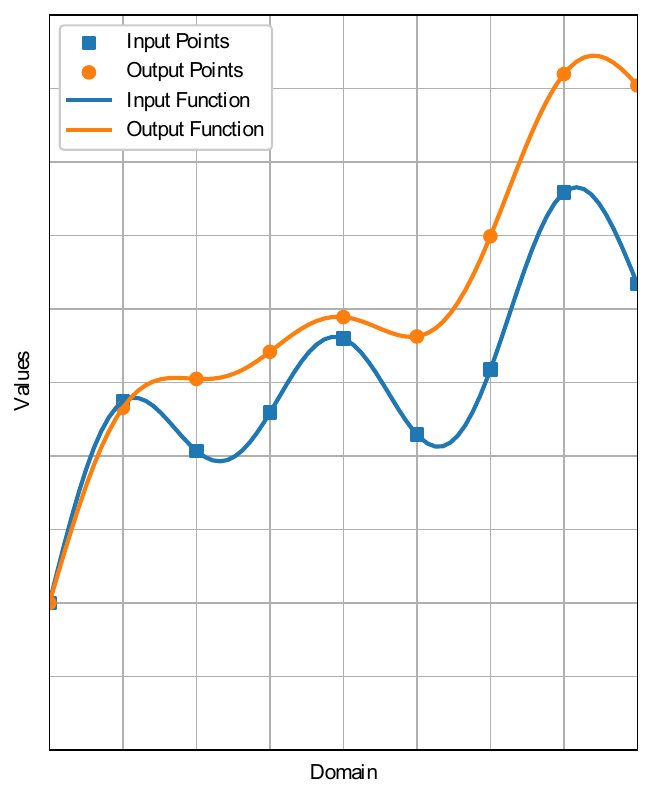}
        \caption{NO maps between functions on continuous domains, even when training data is on a fixed grid.}
    \end{subfigure}%
    \hfill
    \begin{subfigure}{0.32\textwidth}
        \centering
        \includegraphics[width=0.9\textwidth]{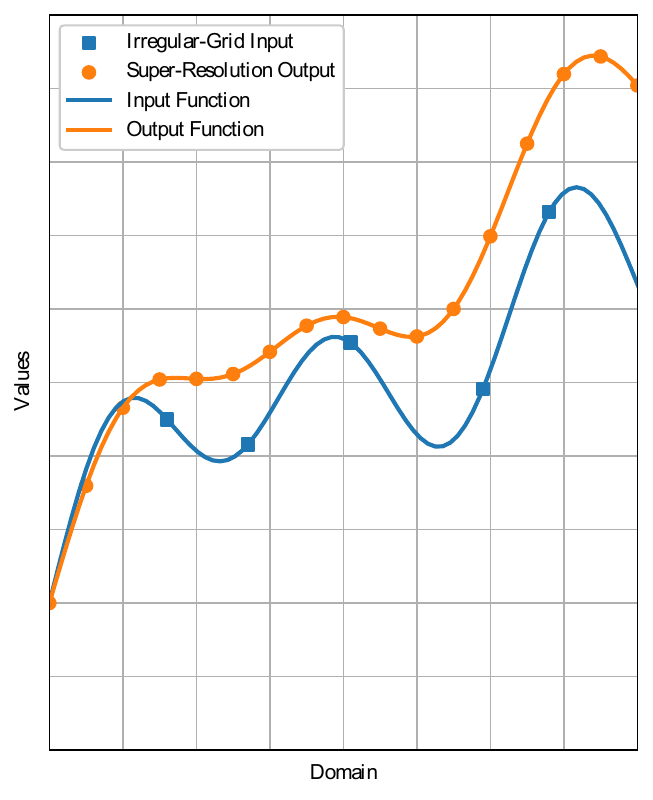}
        \caption{NO maps between functions, so it accepts inputs outside the training grid, and can do super-resolution.}
    \end{subfigure}%
    \caption{Comparison of Neural Networks (NNs) with Neural Operators (NOs). }\label{fig:nnvsno}
\end{figure}

Neural Operators learn operators, which are mappings between two functions defined on continuous domains, and thus can predict at any point in the continuous domain. In the supervised learning setting, the input and output functions are discretized to obtain the training data.  For instance, in the case of weather forecasting, the input consists of variables such as temperature and wind on the earth’s surface and atmosphere, while the output is the values of the same variables at a future time. Here, although the variables are defined on the continuous space-time domain, the training data is only available at a certain discretization, e.g., in the ERA-5 reanalysis dataset, it is 25km and hourly intervals~\cite{hersbach2020era5}. The goal of Neural Operators is to make predictions on continuous domains (such as the Earth's surface and atmosphere for weather forecasting), even though the available training data is only at discrete points.

\begin{figure}[t]
    \centering
    \begin{subfigure}{\textwidth}
        \centering
        \includegraphics[width=0.85\textwidth]{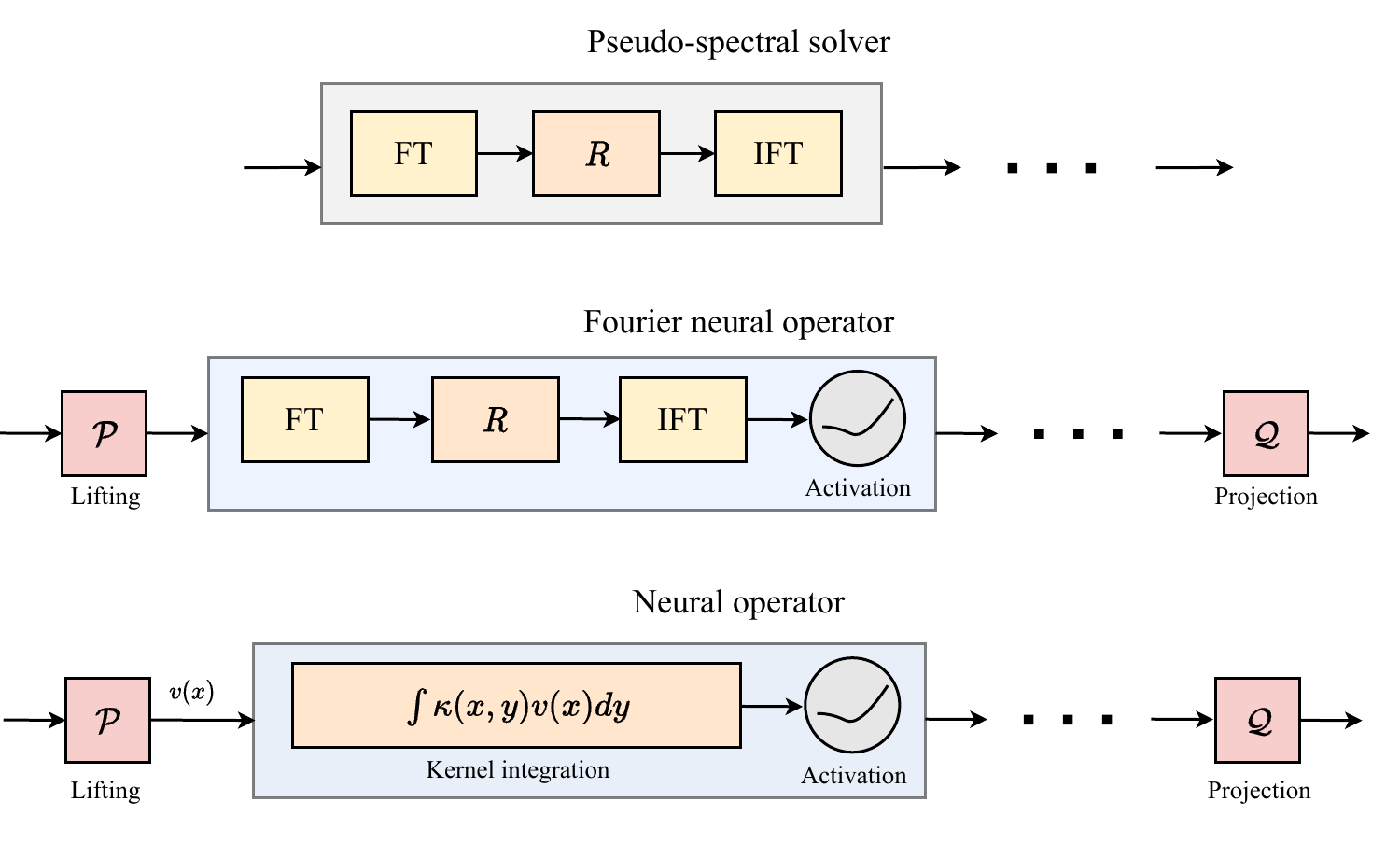}
    \end{subfigure}
    \caption{Diagram comparing pseudo-spectral solver, Fourier Neural Operator (FNO), and the general Neural Operator architecture. FT and IFT refer to Fourier and Inverse Fourier Transforms.  In general, lifting and projection operators $\mathcal{P}$, $\mathcal{Q}$ can be non-linear. Pseudo-spectral solvers are popular numerical solvers for fluid dynamics where the Fourier basis is utilized, and operations are iteratively carried out, as shown. The Fourier Neural Operator (FNO) is inspired by the pseudo-spectral solver, but has a non-linear representation that is learned. FNO is a special case of the Neural-Operator framework, shown in the last row, where the kernel integration can be carried out through different methods, e.g., direct discretization or through Fourier transform. }
    \label{fig:architectures}
\end{figure}

\textbf{Design of a Neural Operator:}
Recall that standard neural networks consist of a series of blocks that perform linear functions, e.g., fully connected or convolutional layers, followed by non-linear activations such as Rectified Linear Units (ReLU). We follow the same philosophy with Neural Operators~\cite{kovachki2023neural}, i.e., designing linear blocks followed by non-linear transformations, with the difference being that linear blocks in Neural Operators are realized by integral operators instead of linear functions in fixed dimensions. 
The linear integral operator is given by 
 \begin{equation}\int \kappa(x,y)a(y)dy\approx\sum_i^N\kappa(x,y_i)a(y_i)\Delta y_i,\label{eqn:kappa}\end{equation} where $a(\cdot)$ is the input function to the operator block, and $\kappa(x,y)$ denotes a learnable kernel between any two points $x$ and $y$ in the  output and input domains respectively. A Neural Operator consists of linear integral operator blocks given by \eqref{eqn:kappa}, followed by pointwise non-linear activations, e.g., Gaussian Error Linear Units (GeLU)~\citep{li2020neural}, and ReLU is usually avoided due to its non-smoothness~\cite{nair2010rectified}. See Fig.~\ref{fig:architectures}.
 
Note that the query point $x$ in the output domain in  \eqref{eqn:kappa} need not be limited to the discrete grid of the training data and can be any point in the continuous domain. Similarly, the input discretization, i.e., grid points $y_i$, can be chosen arbitrarily on which the input function $a(\cdot)$ is specified, and as the discretization becomes finer ($N$ grows larger), this approximation becomes more accurate. 
 Neural Operators can thus take input functions at any discretization, and be queried for predicting solutions at any point in the domain, even if the query point is outside the training grid. We prove that Neural Operators possess a property known as discretization convergence since it converges to a unique operator in the limit of mesh refinement~\cite{kovachki2023neural}. Neural networks do not have such guarantees, even when augmented with interpolation methods such as bilinear interpolation.
This is the fundamental difference between Neural Operators and neural networks, which are limited to fixed dimensions.  See Fig.~\ref{fig:nnvsno}.

\textbf{Zero-shot super-resolution and evaluation:} As a direct consequence of the discretization convergence property of Neural Operators, we obtain the following two properties: i) zero-shot super-resolution, where a trained Neural Operator can receive an input sample at a given resolution and output a prediction at a higher resolution than what is seen during training and ii) zero-shot super-evaluation, where a trained Neural Operator can be evaluated on a new, finer discretization, than what is seen during training, and produce a solution on that new resolution, thus potentially resolving higher frequency details. 

\textbf{Parametrization of the kernel:} While there are several choices to parameterize~\eqref{eqn:kappa}, as outlined below, the choice of non-linear activation is standard, e.g., Gaussian Error Linear Units (GeLU)~\citep{li2020neural}. In addition, Neural Operators also include pointwise lifting and projection operations. This is especially important when the input and output functions are scalar-valued or have a low-dimensional range space. If the kernel is learned in such low dimensions, it has limited expressivity, while the lifting operation expands it into arbitrary channels, enriching the model capacity and expressivity \cite{lanthaler2023nonlocal}. The lifting and projection operations are pointwise and can be linear or non-linear; however,  non-linear operations are, in general, more expressive~\cite{lanthaler2023nonlinear}. Given an architecture and training data, Neural Operators can be trained end-to-end using standard algorithms such as SGD or Adam~\citep{kingma2014adam}, similar to neural networks.

{\bf  Architectures: }Many architectures have been proposed to parameterize the linear integral operation in  \eqref{eqn:kappa}, some drawing inspiration directly from existing numerical methods. For instance, inspired by the classical Petrov-Galerkin method~\cite{j2005introduction},   the kernel $\kappa(x,y)$ in \eqref{eqn:kappa} is constrained to be separable, yielding the DeepONet model~\cite{Lu_2021}. While the original DeepONet model is constrained to be on a fixed input grid, later extensions of DeepOnet \cite{lanthaler2022error, lu2022comprehensive} remove this limitation, which preserves discretization convergence, and thus is a special case of Neural Operators.
Graph Neural Operators (GNO)~\cite{li2020neural} are architectures that are operator extensions of graph neural networks~\citep{battaglia2016interaction} and support the kernel integration on a fixed-radius ball. 
For problems where the ground-truth operator is non-local, GNO has limited expressivity when the radius is small since it cannot capture the global effects due to a limited receptive field. It becomes computationally expensive if we increase the receptive field and set a large radius for the neighborhood. To overcome this and to fit a non-local operator, hierarchical or multi-level extensions have been proposed in \cite{li2020multipole}, inspired by the multipole decomposition of the kernel and multigrid iterative solvers.

The Fourier Neural Operator (FNO)~\cite{li2020fourier} is an alternative approach to obtaining a global operator. In FNO, the kernel integration in \eqref{eqn:kappa} is solved through element-wise multiplication in the Fourier domain with learned coefficients and inverse Fourier transform back to the original domain. The FNO approach is inspired by the pseudo-spectral solvers \cite{trefethen2000spectral} (Fig.~\ref{fig:architectures}) for which the Fourier basis is utilized, and operations are iteratively carried, as shown in Fig.\ref{fig:architectures}. FNO  is inspired by this approach, but has non-linear components to enhance its expressivity. FNO emulates the pseudo-spectral solvers efficiently~\cite{kovachki2021onuniversal}. Moreover, Fourier domain learning is a good inductive bias in many domains, such as fluid dynamics.
For example, it is well known that, in the viscous Burgers' equation, mid-frequency modes are transported to the high frequencies and are subsequently killed off by dissipation. Similar but more complicated dynamics are exhibited in the Navier-Strokes and Euler's equations~\cite{temam2016navier}. By learning directly in Fourier space, the FNO allows us to capture such dynamics efficiently. Extensive theoretical analysis has been developed for Neural Operators in recent years. It has been shown that Neural Operators, in particular FNOs, are universal approximators of continuous operators~\cite{kovachki2023neural,kovachki2021onuniversal}, generalizing the classical results~\cite{chen1995universal} to modern deep learning.

FNO is computationally efficient for capturing global dependencies when the input is limited to a regular grid since the Fourier transform can be computed efficiently through the Fast Fourier Transform (FFT) with quasi-linear complexity. For irregular grids, however, computing the Fourier transform has quadratic complexity, which scales poorly to large problems.  Such a Fourier transform on an irregular grid is proposed in Vandermonde Neural Operators~\citep{lingsch2023vandermonde}.
To enhance computational efficiency, recent methods instead propose deformation~\cite{geoFNO}  or graph kernels~\cite{ginopaper} to transform the irregular physical grid to a regular latent grid on which FFT is applied. Further developments include decomposing the kernel integration in the spatial domain, where the grid is often irregular, while keeping a regular grid in the temporal domain, e.g., for seismic monitoring. Such methods apply GNOs on the spatial domain and FNOs on the temporal domain, resulting in efficient architectures~\citep{sun2023next}. Other developments include tensorized Fourier Neural Operators~\cite{kossaifi2022multi} and  U-shaped Neural Operator~\cite{uno} to efficiently parameterize the Fourier domain. 

Moreover, recent works propose Convolutional Neural Operators for in-place convolutions~\citep{cno} and thus extend CNNs to operator learning. It is further suggested in \cite{kovachki2023neural} that the linear integral transform $\kappa$ in \eqref{eqn:kappa} can be extended to non-linear integral transforms, i.e., $\kappa(x,y,a)$, where the kernel can vary with the input function $a(\cdot)$. One example is the transformer architecture~\cite{vaswani2017attention}. For the special case of a fixed grid with shared input and output domain,  
it has the form 
$ \kappa(x,y,a) := \exp \left< \mathbf{W}^Q a(x), \mathbf{W}^K  a(y) \, \right>/\sqrt{m} $
and the integration is defined as 
\begin{equation}\nonumber
\int \mathcal{S} \left( \kappa\left(x,y,a\right) \, \right) \mathbf{W}^V a(y) dy = 
\int \mathcal{S} \left(
    \frac{\exp\left< \mathbf{W}^Q a(x), \mathbf{W}^K a(y) \, \right>}{\sqrt{m}}\right)
     \mathbf{W}^V a(y) dy\label{eqn:transformer}
\end{equation}
where $\mathcal{S}$ is the softmax function and $\mathbf{W}^Q, \mathbf{W}^K, \mathbf{W}^V$ are the query, key, and value matrices~\citep{cao2021choose}.   While the standard transformer in \eqref{eqn:transformer} is limited to a fixed grid, various attempts have been made to turn it into an operator and allow prediction on irregular grids, e.g., the Operator Transformer (OFormer)~\citep{li2022transformer}, Mesh-Independent Neural Operator (MINO)~\citep{lee2022mesh}, and the General Neural Operator Transformer (GNOT)~\citep{hao2023gnot}.
However, these approaches become computationally intractable since the attention mechanism scales quadratically with the input resolution.
To alleviate this, in vision transformers~\cite{dosovitskiy2020image,AFNO}, 
patches
are used to reduce dimensionality, but the choice of fixed-size patches limits the model to a fixed resolution, and does not yield an operator. It remains an open problem to develop scalable and expressive transformer architectures for operator learning.

 {\bf Implicit Neural Representation (PINN and NeRF): }The concept of using neural networks to represent continuous functions, known as implicit neural networks, has a rich history~\cite{lagaris1998artificial}. One example is when the solution function of a single instance of PDE is represented through a neural network, known as the Physics-Informed Neural Networks (PINN)~\cite{karniadakis2021physics, sirignano2018dgm,yu2018deep,du2021evolutional}. Similarly, in computer vision and graphics, neural radiance fields (NeRF)~\cite{mildenhall2021nerf,sitzmann2020implicit,jeong2022perfception} have been used to represent 3D scenes. Note that in both these cases, the implicit neural network is used to represent a single function, e.g., a single velocity field or a 3D scene. In contrast, Neural Operators can generate the output functions for arbitrary input functions, since they learn the mapping between function spaces. Thus, fundamentally, Neural Operators are a strict generalization of implicit neural networks: from representing a single function to learning operators. In other words, Neural Operators can be viewed as conditional neural fields, conditioned on different input functions. With this viewpoint, there have been several other attempts to extend implicit neural networks to handle multiple instances by additionally learning a parameterization over the space of input functions. However, in general, this mapping is not well posed since one function can be potentially mapped to multiple network parameterizations. To overcome this, recent works use a reduced-order model that maps each input function to a small number of latent parameters (codes) and then feeds it as input to a neural field, similar to an encoder-decoder model~\cite{chen2022crom, serrano2023operator}. In contrast, a Neural Operator is a more direct way to handle multiple instances of input and output functions through operator learning.

{\bf Physics-Informed Neural Operator (PINO): }
PINNs and variants~\cite{fang2023learning} have been effective in solving steady-state PDEs~\cite{raissi2019physics,han2018solving,smith2020eikonet,gao2021phygeonet}. However, the optimization landscape of PINN becomes challenging on problems such as time-varying PDEs, and the solution cannot be easily found using standard gradient-based methods~\cite{pinn-fail}. This issue can be alleviated by extending PINN to a Physics-Informed Neural Operator (PINO)~\cite{pino,wang2021learning,goswami2022physicsinformed}, which incorporates training data in addition to PDE information, making the optimization landscape more tractable and enables us to learn the solution operator of complex time-varying PDEs.

Moreover, PINO has superior generalization and extrapolation capabilities, and reduced training data requirements, when compared to purely data-driven Neural Operators. We can further improve the test-time accuracy of PINO,  by fine-tuning the model to minimize the PDE loss function, on the given PDE instance at test time. This fine-tuning step is identical to the PINN optimization process, but uses the initialization from a pre-trained PINO model and finetunes it further, instead of optimization from a random initialization carried out in PINN. Thus, it can overcome the optimization difficulties in PINN.

Further, in PINO~\cite{pino}, we can combine low-resolution training data with high-resolution physics constraints to learn a good approximation of the underlying operator and carry out accurate zero-shot super-resolution. Fig~\ref{fig:spectrum} presents results after training with the FNO backbone on low-resolution data, i.e., $64\times64$, and further fine-tuned using PDE loss at higher resolution (PINO) and tested on high-resolution data, $256\times256$.  We believe that having such multi-resolution losses is crucial to obtaining high-fidelity learned solvers that can resolve fine-scale features.

An alternative set of techniques is motivated by the Monte Carlo methods, where we can leverage the probabilistic representation of PDEs to obtain suitable objectives for deep learning. Specifically, reformulations of PDEs based on the Feynman-Kac formula and backward stochastic differential equations have been successfully employed to tackle high-dimensional parabolic and elliptic PDEs, see, e.g.,~\cite{han2018solving,berner2020numerically,han2020derivative,beck2021solving,richter2022robust}.
Recently, such ideas have been combined with Neural Operators, promising efficient, derivative-free alternatives to Physics-Informed Neural Operators~\cite{zhang2023monte}.

{\bf Generative Neural Operators: }We have so far discussed Neural Operators for approximating deterministic mappings between function spaces. There are also extensions to learning probabilistic mappings on function spaces.  These are either based on generative adversarial networks (GAN)~\cite{gano}, diffusion models~\cite{dfno}, or variational autoencoders~\cite{seidman2023variational} that are extended to function spaces. In particular, the Diffusion Neural Operator learns a score operator that takes a Gaussian random field as input and can generate samples at any resolution specified. This has been applied to problems such as volcanic and seismic activities, statistics of Navier Stokes equations, and resolution-free visual sensors~\cite{gano,dfno,shi2023broadband}.

{\bf Applications: }Neural Operators have yielded impressive speedups over traditional numerical solvers in a wide range of domains. They can do medium-range weather forecasting tens of thousands of times faster than current numerical weather models, and were the first data-driven approach that resulted in accurate high-resolution (0.25 degrees) weather forecasting~\cite{pathak2022fourcastnet}. This speedup also enables accurate risk assessment of extreme weather events such as hurricanes and heat waves since we can create many ensemble members. An improved version based on the spherical Fourier Neural Operator~\cite{bonev2023spherical} is now running on an experimental basis alongside traditional forecasting systems at the European Centre for Medium-Range Weather Forecasts (ECMWF)~\cite{url:ecmwf}. Similarly, for the application of Carbon Capture and Storage (CCS) for climate-change mitigation, the nested FNO model is hundreds of thousands of times faster, enabling large-scale assessment of reservoirs for $\textrm{CO}_2$ storage~\cite{wen2023real}. We present a rigorous probabilistic assessment for maximum pressure
buildup and carbon-dioxide plume footprint, with a runtime of only 2.8 seconds with our FNO model, while this would have taken nearly two years with the numerical simulators. Other applications include simulations of fluid dynamics~\cite{grady2022towards, renn2023forecasting, liz2022fourier}, 3D industrial-scale automotive aerodynamics~\cite{ginopaper}, 3D 
dynamic urban microclimate~\cite{peng2023fourier},
material deformation~\cite{liu2022learning, rashid2022learning,geoFNO},
computational lithography~\cite{liu2022adversarial, yang2022generic},
photo-acoustics~\cite{guan2021fourier},
and electromagnetic fields~\cite{gu2022neurolight, gopakumar2023fourier}. Further, Neural Operators have been used for learning long-term statistical properties such as attractors~\cite{li2022learning,lippe2023pde} in chaotic systems, and detecting tipping points in non-stationary systems~\cite{liu2023tipping}. Additionally, Neural Operators based on generative adversarial networks~\cite {gano} or diffusion models~\cite{dfno} in function spaces can learn natural phenomena that are stochastic, e.g., volcanic activities and stochastic differential equations~\cite{salvi2022neural}. Also, researchers have explored 
building hybrid solvers using Neural Operators with numerical solvers~\cite{ngom2021fourier}.

Neural Operators are also effective for inverse problems, where we aim to find a parameter from noisy observations, and in many cases, the forward model is non-invertible or has a poorly conditioned inverse. Examples include inverse design, optimization, control~\cite{shi2022machine}, and risk assessment. 
Since these models offer fast inference and can be efficiently differentiated, they can be used for either sampling the Bayesian solution to an inverse problem through MCMC methods \cite{cotter2012mcmc}, or for obtaining a point estimate by optimizing a regularized functional \cite{hinze2008optimization}. An invertible Neural Operator framework~\cite{kaltenbach2023semi} has been introduced for such problems. In inverse design, the goal is to optimize design over a given set of parameters, given a  forward model, and provide guidance for iterative design improvements or generate optimized designs from scratch. The Neural Operator model is able to generate an optimized design for a novel medical catheter that reduces bacterial contamination by two orders of magnitude~\cite{zhou2023ai}. The Neural Operator model can accurately simulate the bacterial density in fluid flow and optimize design shapes within the catheter to prevent bacteria from swimming upstream into the human body.  Neural Operators are also effective for other inverse problems, e.g.,  in seismology, given the seismic waves and earthquakes observed on the subsurface of Earth, we need to infer the structure of the subsurface causing the specific seismic observation on the surface~\cite{yang2021seismic,sun2022accelerating,yang2023rapid,yin2023solving}. Thus, Neural Operator methods enable inverse problems at considerably higher fidelity than prior methods and are significantly faster. Neural Operators are having a transformative impact in a wide range of scientific domains.

{\bf Accelerated Software Implementation and Benchmarks: }Efficient software and hardware implementation are crucial to successfully applying Neural Operators to practical problems. We provide a reference implementation for various Neural Operators in PyTorch and examples to get started~\cite{url:no}. In addition to libraries, standardized benchmarks are crucial for further development. However, this has been a challenge due to the disparate nature of different systems across scientific fields. For instance, the choice of PDE parameters (e.g., forcing functions, domain size, resolution of training data, etc.)  can significantly impact its dynamics. There are two main categories of PDE benchmarks: (1) broad, unified benchmarks for machine learning in physical systems~\cite{otness2021extensible,takamoto2022pdebench,gupta2022towards,hao2023pinnacle} and (2) task or domain-specific benchmarks~\cite{huang2021large,ren2023superbench,wang2004image,hassan2023bubbleml,dulny2023dynabench}. 

In order to aid the further development of Neural Operators, it is critical to construct more universal benchmarks that can correctly assess operator-learning capabilities.  In our opinion, many current benchmarks do not meet these requirements, and many are limited to training and evaluation to a single resolution~\cite{gupta2022towards} or do not provide well-resolved training data for operator learning to be effective in the purely data-driven setting~\cite{gupta2022towards}. In addition to learning metrics, the benchmarks also need to incorporate hardware performance metrics that are typically used in HPC studies~\cite{10.1145/3592979.3593412} and explore the benefits of low-precision learning~\cite{white2023speeding}.

\section{Challenges and Outlook}

We have shown that Neural Operators are a principled AI approach for modeling multi-scale processes in various scientific domains, e.g., fluid dynamics, wave propagation, and material properties.  Multi-scale processes span across many scales of interactions, where microscopic interactions impact macroscopic behavior. Current numerical methods are too expensive to simulate such processes faithfully since they require iterating on a fine grid. Neural Operators achieve significant speedups over current methods by learning non-linear feature transformations and overcoming the need for a fine grid. In addition, they are naturally suited for inverse problems since they are differentiable, and we can directly invert the trained forward model through gradient-based optimization. In contrast, traditional simulations are non-differentiable and require expensive Markov Chain Monte Carlo (MCMC) for inverse problems. Neural Operators learn mappings between function spaces and, as such, can be queried at any discretization and not just limited to the training resolution. Thus, Neural Operators are fundamentally different from neural networks, which are limited to a fixed resolution or dimension.    
 
There are, however,  several open problems and limitations with the current approaches. For instance, there are practical challenges in dealing with massive datasets for training Neural Operators. In contrast to many traditional deep learning tasks in computer vision or language,  in scientific computing, each data point can itself be massive. For example, in weather forecasting, each data point constitutes hundreds of variables defined over the entire globe at high resolution. The massive memory and bandwidth requirements for data storage and movement bring many challenges, such as developing efficient hardware and software frameworks that can handle compressed formats~\cite{url:zarr}, and also algorithmic approaches such as multi-grid patching~\cite{kossaifi2022multi} and other parallelization approaches. Moreover, there are still many challenges in ensuring good learning and generalization of Neural Operators. For instance, data-driven models are not guaranteed to have physical validity, which may be important in certain domains, e.g., climate modeling. Moreover, they may have challenges in capturing rare events due to the lack of training data representing such events. For instance, in FourCastNet and other AI-based weather models, extreme weather events such as hurricanes are predicted reasonably well, but differ in intensity when compared with the ground truth~\cite{url:intensity-ecmwf}. 

The generalization challenges can be alleviated to some extent using hybrid operator learning that combines data and physics constraints. However, deep fundamental challenges remain. Gradients from PDE constraints can be calculated to different levels of fidelity with increasing costs in different methods, e.g., autograd, Fourier continuation~\cite{maust2022fourier}, and finite difference~\cite{pino}. The use of multi-fidelity approaches to achieve an optimal tradeoff remains an open problem. In chaotic system forecasting, ensuring the accuracy of learned system evolution over long time scales is another open problem~\cite{li2022learning, lippe2023pde}, since in chaotic systems, small forecasting errors compound over time. For such problems, long-term conservation laws and other physics constraints need to be incorporated into operator learning~\citep{ben2023rise}.
Moreover, overcoming optimization challenges in data-limited regimes when data points are expensive requires new active learning methods for cost-efficient data collection. In high-dimensional settings, there is a need for novel integrations, e.g., Monte-Carlo sampling~\citep{berner2020numerically,zhang2023monte} integrated with methods already explored in PINNs, e.g., adaptive sampling. Moreover, developing efficient uncertainty quantification methods for function spaces still remains an open problem~\cite{Psaros_2023}, especially in high dimensions.  Further, we also believe that there is an opportunity to build a more universal foundation model for simulation and modeling, and Neural Operators serve as an ideal  backbone~\cite{subramanian2023towards}, where the data-driven approach can be augmented with the guardrails of physics constraints, such as conservation laws and symmetries,  to allow for faithful extrapolation and generalization. 

Thus, in summary, Neural Operators are enabling more efficient scientific modeling, leading to new inventions and discoveries. By obtaining significant speedups over traditional simulations, we can enable cheaper modeling and scaling to larger systems that would not have been possible before. Due to their differentiability, Neural Operators are also effective on inverse problems such as design, which allows us to create new inventions at a much faster pace than before, while ensuring physical validity.

\backmatter

\bmhead{Acknowledgments} A. Anandkumar is supported by the Bren named professor chair at Caltech. Zongyi Li is supported by the NVIDIA fellowship. Miguel Liu-Schiaffini is supported by the Mellon Mays undergraduate fellowship. We thank Benedikt Jenik for creating Figure 2 and other general discussions.

\bibliography{main}

\end{document}